# What Knowledge is Needed to Solve the RTE5 Textual Entailment Challenge?

**Working Note 39**
Peter Clark (peterc@allenai.org)
June 2010, submitted to arXiv 2018

## Introduction

This Working Note gives a knowledge-oriented analysis of about 20 interesting Recognizing Textual Entailment (RTE) examples, drawn from the 2005 RTE5 competition test set [1]. The analysis ignores shallow statistical matching techniques between T and H, and rather asks: What would it take to reasonably infer that T implies H? What world knowledge would be needed for this task? Although such knowledge-intensive techniques have not had much success in RTE evaluations, ultimately an intelligent system should be expected to know and deploy this kind of world knowledge required to perform this kind of reasoning.

The selected examples are typically ones which our RTE system (called BLUE, [5]) got wrong and ones which require world knowledge to answer. In particular, the analysis covers cases where there was near-perfect lexical overlap between T and H, yet the entailment was NO, i.e., examples that most likely all current RTE systems will have got wrong. A nice example is #341 (page 26), that requires inferring from "a river floods" that "a river overflows its banks". Seems it should be easy, right?

The analysis uses WordNet's glosses as a possible source of world and lexical knowledge (on the assumption the glosses were fully and correctly axiomatized, which they are not), with additional, occasional references to DIRT [2], DART [3], and PowerSet's Factz database [4]. Other sources are of course also possible, but working with concrete, existing resources helps anchor the analysis in real linguistic and reasoning manipulations that people do easily, but are hard for machines.

Each example is presented twice: The full example, and then an abbreviated form, showing just the important parts of the sentences. In many cases, the bulk of the the T sentences are irrelevant, so you can just jump straight to the abbreviated example when reading through this. Of course, deciding which parts of the sentence are relevant and irrelevant is itself a difficult computational task.

Enjoy!

## Index

**#2: NO** (BLUE said YES. This example is interesting as there is **perfect lexical overlap** of H on T, i.e., is a case where lexical overlap is a bad metric to use).

## Full example:

> **T:** Euro MPs have voted overwhelmingly to cut the cost of texting and using the internet on mobiles abroad. The cap for a "roaming" text will fall to 11 euro cents (10p; 14 US cents), from about 29 cents on average today. The EU-wide caps, excluding VAT, will take effect in July. They cover text messages and data roaming services, such as checking e-mails while abroad. The current price cap of 46 euro cents per minute for an outgoing voice call will also fall to 43 cents in July.
>
> **H:** A roaming text cost 46 euro cents.

## Abbreviated example:

> **T:** ...The cap for a "roaming" text will fall...from about 29 cents today....
>
> **H:** A roaming text cost 46 euro cents.

## Analysis:

### Theme: Proving Contradictions of Functional Predicates

This is a good example of the inherent limitations of lexical overlap. The provable contradiction here is:

> **T:** X costs at most Y.
> **H:** X costs Z
>     *and* Y is less than Z.

A solution strategy would be as follows: To refute f(x,y), where f is functional, look for f(x,y') and then show y and y' are inconsistent. Or in this case, look for some constraint C on arg2 of f(x,y), and show y is inconsistent with C. In this case, to refute **H:**cost(X,Z), we find T: cost(X,W), W<Y, Y<Z, and show W cannot equal Z.

### Theme: Commonsense Knowledge

We also need knowledge that

> IF the [price] cap of X is C THEN X must cost less than C.

and

> IF X falls from Y THEN X was at Y originally

We also need to recognize "cap" as a "price (cost) cap". This meaning is very strongly hinted at in the first sentence:

> "MPs have voted...to cut the cost of texting...".

And we know that one way MPs can cut costs is to impose a cap. To use this, we'd also need to relate "texting" to "a text".

### Theme: Metonymy

Note also the metonymy: "The cap for a text" means "The cap for *sending* a text" (you pay to send a text, not to obtain ownership of it).

WordNet tells us:

> cap#n6: ceiling, cap -- (an upper limit on what is allowed; "they established a cap for prices")

**#4 YES** (BLUE said UNKNOWN. This is interesting as it is a very complex example)

## Full example:

> **T:** A Spanish man who has admitted killing a Welsh couple, has expressed remorse for his actions and apologised to their son in his murder trial. Brian and Tina Johnson were bludgeoned to death with a hammer on Fuerteventura in the Canary Islands in July 2006. Juan Carmelo Santana told the jury at Las Palmas criminal court that he blamed the drink and drugs he had taken for the attacks. The jury is expected to begin its deliberations later.
>
> **H:** Mr. and Mrs. Johnson were killed by Santana.

## Abbreviated example:

> **T:** A Spanish man who has admitted killing a Welsh couple...Brian and Tina Johnson were bludgeoned to death....Juan Carmelo Santana told the jury at Las Palmas criminal court that he blamed the drink and drugs he had taken for the attacks.
>
> **H:** Mr. and Mrs. Johnson were killed by Santana.

## Analysis:

**Theme: Pragmatics, coreference**

The first challenge is resolving the coreference between "a Welsh couple" (sentence 1) and "Brian and Tina Johnson" (sentence 2). This is an example of a very common pragmatic structure in articles of:

> *introduction*
> *elaboration*

When *elaboration* occurs, we have to align the elements in it with *introduction*, even if there is no definite reference. We would say there is an "elaboration" rhetorical relation between *introduction* and *elaboration*.

For example:

> "A man fell to his death. Joe slipped while climbing and tumbled..."

In principle these two sentences could be talking about different events, but pragmatic constraints suggest that they are the same event. Thus the first sentence creates some initial objects in the scenario (man, fall, death) and objects in the second sentence (Joe, tumbled) are aligned with them. To make matters more complicated, it is considered good literary style to use different words for the same thing to avoid boring the reader. Thus there is a lot of coreference to sort out.

Proper names are a kind of definite reference, either to someone (or thing or organization) we already know, either from background knowledge ("Obama") or introduced elsewhere (usually earlier) in the text. When a new person is introduced (i.e., the name is not intended as a reference), there is normally some information given for the name to refer to, e.g., "Joe, a resident of London, ...". (We can view this latter case as an example of forward reference, c.f., "A resident of London, Joe, ..." which would be the more traditional backward reference). "Naked names", i.e., names without any referent, are unusual in text. (Sometimes a "naked name" is used for rhetorical effect, e.g., a story or newpaper article might begin: "Sue Ashley could not pay her bills. ...")

In this RTE example, we have the same thing:

> "A man...killed a Welsh couple.
> Brian and Tina Johnson were bludgeoned to death with a hammer..."

Although there is no explicit reference in the second sentence, it clearly refers back to the first.

[ Note that if we add "also" it changes the rhetorical structure:

> "A man...killed a Welsh couple.
> "Also, Brian and Tina Johnson were bludgeoned to death with a hammer..."

this changes the rhetorical role of the second sentence to be *addition* (new knowledge) rather than *elaboration*, and so the alignment would no longer be justified. It also makes "Brian and Tina Johnson" sound strange, as it is a reference but there is no longer a referent for it. ]

**Theme: Coreference and Commonsense Expectations**

The coreference of "A Spanish man" and "Juan Carmelo Santana". One might infer this weakly by the fact "Juan Carmelo Santana" is a Spanish name. But more importantly, there is a trial going on:

> "A Spanish man...has expressed remorse for his actions and apologised to their son in his murder trial."

Then at the trial:

> "Juan Carmelo Santana told the jury..."

Note we need to know that jury is part of the trial, to identify that "Juan Carmelo Sanatan" is coreferential with "A Spanish man". At this point, Juan could be anyone participating in the trial (judge, defendant, prosecution, defense, or witness; note we again need to know this stereotype of how a trial works). But then:

> "Juan Carmelo Santana told the jury...that he blamed the drink and drugs he had taken for the attacks."

The phrase "he had taken" is quite important. If T had said:

> "Juan Carmelo Santana told the jury...that he blamed the drink and drugs the murderer had taken for the attacks."

then Juan would more likely be the defense attorney (or conceivably the prosecution) rather than defendant. However, there's a common stereotype being conveyed here: Someone does something bad (murder is bad) and tries to make an excuse for it (blame something/someone else). It fits nicely to the story, so this is how we understand it. There are other hints: "expressing remorse" and "blaming the drink and drugs" seem to go together.

**Theme: Modal Statements**

"A...man..admitted killing..." → "A man killed..."

Basic linguistic analysis should get this.

**Theme: Coreference and Commonsense Knowledge**

"A Spanish man...has expressed remorse for his actions and apologised to their son in ***his*** murder trial."

"***his***" in "his murder trial" refers "the Spanish man", not the nearest preceding male in the dialog "their son". This takes some quite sophisticated understanding to get: You don't normally try a deceased person; the murderer has already been identified earlier.

**#6: YES** (BLUE said YES). This is an easy example, computable by simple phrase subsumption.

## Full example:

| |
|---|
| **T:** Rain is pelting down on Doña Porcela's treatment room in Puerto Cabezas, the main town on Nicaragua's Northern Caribbean coast. The room is barren except for a few plastic chairs, a wooden table and some old plastic bottles balanced precariously on timber beams. Doña Porcela is a respected traditional healer here and the bottles are filled with her secret medicinal potions. Her patient today is a teenage girl asleep on a piece of cardboard, serving as a mattress on the dirt floor. |
| **H:** Doña Porcela is a healer. |

## Abbreviated example:

| |
|---|
| **T:** Doña Porcela is a respected traditional healer. |
| **H:** Doña Porcela is a healer. |

## Analysis

This is very straightforward.

**#7: NO** (BLUE said YES. Another interesting and humorous example with **perfect lexical overlap** of H on T but no entailment)

## Full example:

**T:** Global technology giant IBM is in talks to buy Sun Microsystems in a deal that would expand its server market share, the Wall Street Journal reported Wednesday. IBM may pay as much as US$6.5 billion in cash for Sun, the newspaper reported on its Web site, without naming its sources. That amount of money would be nearly double Sun's closing share price on Tuesday of $4.97 per share. The report cautioned that while the two companies are holding discussions, a transaction may not occur.

**H:** The price of Sun Microsystems is $4.97.

## Abbreviated example:

**T:** ...Sun's closing share price on Tuesday of $4.97 per share....

**H:** The price of Sun Microsystems is $4.97.

**Theme: Metonymy**

This is a cute example! One could reasonably argue that the answer is actually YES, where "The price of Sun Microsystems" is metonymy for "The price of Sun Microsystems' shares". We often say things, in context, like "Boeing was up to $50.43". It's only here because there's discussion of the value of Sun Microsystems that $4.97 looks silly. The abbreviated version (above) is quite plausible. I personally think it's reasonable to assign "YES" to this one.

**#8: YES** (BLUE said YES). A simple example of apposition.

## Full example:

| | |
|---|---|
| **T:** | DEFENDER Zdenek Grygera scored on a header in added time to earn 10-man Juventus a 1-1 draw with Inter Milan on Saturday, although it may have not been enough to prevent Inter from clinching its fourth successive Serie A title. Teenager Mario Balotelli scored in the 64th minute for Inter and Juventus midfielder Tiago Mendes was shown a red card in the 76th. "To go down a goal, lose a man, and then be able to draw, satisfies me," Juventus coach Claudio Ranieri said. Inter remained 10 points ahead of Juventus with six rounds remaining. |
| **H:** | Claudio Ranieri is the Juventus' coach. |

## Abbreviated example:

| | |
|---|---|
| **T:** | Juventus coach Claudio Ranieri |
| **H:** | Claudio Ranieri is the Juventus' coach. |

## Analysis

This is a simple, straightforward example.

**#9: YES** (BLUE said UNKNOWN)

## Full example:

> **T:** India's economy has grown by more than 8 per cent annually since 2003 and hit $4 trillion (based on purchasing power parity) by the end of last year - more than double that of the whole of Africa. The country now has the sort of budget, foreign exchange reserves, transport infrastructure, human resources and stable political environment that are the envy of most sub-Saharan countries. Yet its child malnourishment levels are worse than Ethiopia's and on a par with those of Eritrea and Burkina Faso.
>
> **H:** India, Eritrea and Burkina Faso have the same rate of child malnourishment.

## Abbreviated example:

> **T:** India's economy....The country....Yet its child malnourishment levels are...on a par with those of Eritrea and Burkina Faso.
>
> **H:** India, Eritrea and Burkina Faso have the same rate of child malnourishment.

## Analysis

**Theme: Subtle semantics**

Strictly H isn't entailed for 2 reasons:

1. "rate" and "level" aren't quite synonyms, as

> "level" (level#n1) is a kind of "magnitude" (magnitude#n1) [WordNet]

but

> "rate" (rate#n1) is a kind of magnitude_relation#n1 (a relation between magnitudes)

In fact, the H phrase "rate of child malnourishment" doesn't quite make sense - at least, it's not temporal (# malnurished / unit of time), it's # malnurished / unit of population. Which can be viewed as a "level" - really a "level" can be any quantity, a rate can be a level. One might argue that "rate" isa "level" -- but for this entailment we need the reverse, that "level" (T) is a kind of "rate" (H). I think overall that (senses of) {rate,level,amount} are synonyms, although WordNet says otherwise.

2. "on a par with" doesn't exactly mean "same", it only means "similar":

Interestingly, DIRT does have the paraphrase:

> IF X is on a par with Y THEN X is about the same as Y

This is pretty darn good, especially it (desirably) includes "about". However, DIRT (incorrectly) has "same" as a noun, which doesn't match with our parse of the sentence where "same" is an adjective (applied to an implicit object) :-(

> DIRT: "is the same as" → "be"(X,same01), "as"(same01,Y).
> BLUE: "is the same as" →

"be"(X,thing01), "as"(thing01,Y), modifier(thing01,same01).

"same" itself is a slippery concept. "same" to what level of precision? No two things are *exactly* the same. Our entailment engine concludes that "about the same" → "the same", as it simply drops "about" (!). This is reasonable for "about", but not for polarity-reversing adjectives such as "not".

**#11: NO** (BLUE said YES, another example with **perfect lexical overlap** but no entailment)

## Full example:

| |
|---|
| **T:** A Soyuz capsule carrying a Russian cosmonaut, an American astronaut and U.S. billionaire tourist Charles Simonyi has docked at the international space station. Russian cosmonaut Gennady Padalka manually guided the capsule to a stop ahead of schedule Saturday two days after blasting off from the Baikonur cosmodrome in Kazakhstan. The crews of the capsule and the station will spend around three hours checking seals before opening the air locks and meeting up face-to-face. |
| H: Charles Simonyi is a Russian cosmonaut. |

## Abbreviated example:

| |
|---|
| **T:** ...a Russian cosmonaut, an American astronaut and U.S. billionaire tourist Charles Simonyi... |
| **H:** Charles Simonyi is a Russian cosmonaut. |

## Analysis

**Theme: Lists and appositives**

There's a need to recognize a list structure here, and realize that the members are distinct. Note that we have to be careful to distinguish a list from an apositive, e.g.,:

> "Einstein, [who is] an Austrian citizen and brilliant scientist, discovered relativity..."

Our commonsense knowledge that Russian and American are mutually exclusive, as are astronaut and tourist, is useful here.

**#14 YES** (BLUE said UNKNOWN. This is a highly complex example)

## Full example:

| |
|---|
| **T:** Three major bombings in less than a week will be causing some anxiety among political leaders in Baghdad and Washington. Last Thursday 10 people were killed by a car bomb at a crowded cattle market in Babel province, south of Baghdad. On Sunday more than 30 died when a suicide bomber riding a motorbike blew himself up at a police academy in the capital. Tuesday's bombing in Abu Ghraib also killed and wounded a large number of people - including journalists and local officials. |
| **H:** In less than a week there were 3 major bombings in Iraq, killing more than 40 people. |

## Abbreviated example:

| |
|---|
| **T:** Three major bombings in less than a week...10 people were killed by a car bomb...south of Baghdad...more than 30 died when a suicide bomber...blew himself up...in the capital...Tuesday's bombing in Abu Ghraib... |
| **H:** In less than a week there were 3 major bombings in Iraq, killing more than 40 people. |

## Analysis

**Theme: Coreference**

There is implicit coreference here: "Three major bombings..." then a sentence describing each one (implicitly, each is one of the three).

**Theme: Arithmetic**

This example requires summing the numbers "10" + "more than 30" to conclude "more than 40"

**Theme: Ellipsis (implicit noun)**

Implicit noun "more than 30" → "more than 30 people"

**Theme: Geographical Reasoning**

H requires concluding that the location of each of the 3 bombings were "in Iraq". This is pretty darn hard to infer, and needs to be inferred from the heavy hints in T rather than direct implication. The fact they "caused anxiety among...leaders in Baghdad and Washington", along with political knowledge that Iraq bombings are (currently) common and a concern, strongly hints the bombings are in Iraq. Specifically on each of the 3 bombings in T some geographic knowledge would also help:

    a. "in Babel province, south of Baghdad" → in Iraq

For this, need to know Baghdad (and thus a province south of Baghdad) is in Iraq

    b. "in the capital" → in Baghdad → in Iraq

    c. in Abu Ghraib → in Iraq

**#16 YES** (BLUE said UNKNOWN)

## Full example:

> **T:** Looking drawn and moving stiffly after three days in the hospital this week, Cardinal Edward M. Egan returned to the pulpit of St. Patrick's Cathedral on Thursday to participate in Holy Week services for the last time as archbishop of the Roman Catholic Archdiocese of New York. A little more than a week before his retirement, Cardinal Egan, 77, was stricken with severe stomach pains on Saturday night and driven to St. Vincent's Hospital Manhattan in Greenwich Village. He was released on Tuesday after treatment for a gastrointestinal virus, with an unspecified appointment to come back for pacemaker implantation surgery, and with instructions from doctors to rest.
>
> **H:** Edward M. Egan is the New York Catholic Archbishop.

## Abbreviated example:

> **T:** ...Edward M. Egan returned ...to participate ...for the last time as archbishop of the Roman Catholic Archdiocese of New York....
>
> **H:** Edward M. Egan is the New York Catholic Archbishop.

## Analysis

**Theme: Different syntactic ways of rendering a complex set of relationships**

Somehow we have to equate the fairly different names:

>  "archbishop of the Roman Catholic Archdiocese of New York"

>  "New York Catholic Archbishop"

**Theme: Modal Statements**

>  "X returns to <act>" → "X <acts>"

>  "X participates as Y" →
>       "X is Y"  (here, "participates as archbishop" → "is archbishop")

This isn't universally true but is common. The equivalence here is because "archbishop" is being treated (in places) as a role. In other cases, though, "participate as" might mean "stand in as" or "pretend to be".

**#17: YES** (BLUE said UNKNOWN)

## Full example:

> **T:** Although it won't discuss it, the Chinese government frequently blocks foreign Web sites that it deems to have objectionable content. Access to YouTube was also blocked in 2007, and 2008 during high-profile government meetings and following last year's unrest in Tibet. YouTube has been blocked at least twice in the last month, according to Herdict Web, a censorship tracking system run by Harvard University's Berkman Center for Internet and Society. The first block came on March 4, coinciding with the 50th anniversary of the Tibetan Uprising on March 10. The current block began on March 23.
>
> **H:** The ban is not the first ban for YouTube in China.

## Abbreviated example:

> **T:** ...the Chinese government frequently blocks...Access to YouTube was also blocked in 2007, and 2008...YouTube has been blocked at least twice in the last month....The first block came on March 4...The current block began on March 23.
>
> **H:** The ban is not the first ban for YouTube in China.

## Analysis

**Theme: Coreference between T and H**

The coreferences between "the ban" and "the first ban" (H) and elements of T are horrible. "the ban" (H) presumably refers to "The currrent block (right at the end of T). "the first" (H) is horrible to understand; strictly, it requires representing the chronology (timeline) of blocks imposed by China.

**Theme: Synonymy, and the use of presuppositions for coercion**

"block" and "ban" are not related in WordNet, and in fact they are different concepts (although should be treated as coreferential for the purposes of this example). Strictly a ban is a legal degree, while a block is an action. China could block YouTube without officially banning it. Some real-world reasoning might infer that if X is blocked, then X can't be accessed, which is a means of imposing a ban.

H contains an interesting mix of presupposition and query. "The ban..." in H presupposes there is a ban, rather than asks if there was one. Instead the focus of the query in H is whether that ban was the first or not. Thus, maybe the RTE system should similarly assume that there is a ban (and thus coerce "block" to match it even if there's not a direct connection) rather than search for one. Consider that if H had alternatively said "China banned YouTube", then that might be considered UNKNOWN to a reader (as the equivalence of block and ban is now the question, rather than a presupposition).

**Theme: Implicit knowledge**

Also from T: ...Chinese government frequently blocks..." the system needs to assume that all subsequent blocking incidents described in the text are also by the Chinese government (hence China). i.e., the first sentence sets up the context for the rest of the sentences.

**Theme: Grain size/granularity reasoning**

We also need: IF the Chinese government does X THEN China does X

**#20 YES** (BLUE said UNKNOWN)

## Full example:

> **T:** Ian Tomlinson, 47, died last Wednesday after encountering riot police near the Bank of England as he returned home from his work as a newsagent. The footage, released yesterday by The Guardian, reveals a docile-looking Mr Tomlinson walking with his back to police officers at about 7.20pm. One Metropolitan Police officer appears to lunge at him with a baton from behind and he falls to the ground. A bystander had to help him to his feet. Mr Tomlinson is understood to have then walked on to Cornhill where he collapsed. A New York fund manager visiting London shot the footage.
>
> **H:** Ian Tomlinson was a newspaper vendor.

## Abbreviated example:

> **T:** Ian Tomlinson...work[ed] as a newsagent.
>
> **H:** Ian Tomlinson was a newspaper vendor.

## Analysis

**Theme: Definitional knowledge**

"X works as a Y" means (here) "X is a Y" (here Y = newsagent = a role). This requires basic definitional knowledge! WordNet gives us:

>    newsagent#n1: someone who sells newspapers

**Theme: Morphosemantics**

We also need to know that if you sell something, you are the seller (morphosemantics). {seller,vendor} are synonyms in WordNet.

**#36 YES** (BLUE said UNKNOWN. WARNING: This is such an incredibly complex example it might be dangerous to your health...)

## Full example:

| |
|---|
| **T:** The U.S. State Department said Thursday that the United States is working to dissuade North Korea from going ahead with its plan to launch a satellite next month. U.S. officials say the launch is a disguised long-range missile test and would be destabilizing. State Department officials give no credence to the notion that the launch would be for scientific purposes and they say the United States is working diplomatically to try to prevent a North Korean action that they say would be provocative, destabilizing and unhelpful. |
| **H:** The U.S. fears that the North Korean satellite is a ruse to hide the testing of a missile. |

## Abbreviated example:

| |
|---|
| **T:** ...North Korea...its plan to launch a satellite...U.S. officials say the launch is a disguised long-range missile test.. |
| **H:** The U.S. fears that the North Korean satellite is a ruse to hide the testing of a missile. |

## Analysis

**Theme: Grain size/granularity reasoning, or Metonymy**

Need to equate "The U.S. State Department says" (T) with "The U.S says" (H). This might entail knowing that the State Department is a kind of mouthpiece for the US.

**Theme: Modal Statements**

North Korea plans to launch a satellite → the satellite belongs to North Korea → **H:** North Korean satellite

**Theme: Metonymy**

There's metonymy in **H:** "the...satellite is a ruse" really means "the [launch of] the satellite is a ruse".

**Theme: Reasoning with World/Lexical knowledge**

We need to resolve the above metonymy to match the following:

    **T:** "the launch is a disguised...test"
    **H:** "the [launch of] the satellite is a ruse"

WordNet gives us some information:

        disguise#n1 isa hide#n1        [1]
        disguise#n1 isa deception#n1    [2]
        ruse#n1: a deceptive maneuver    [3]

From these there is a partial line of reasoning. Looking at implications of T, and equalities of H, we see:

**T:** "the launch is a disguised...test"
   → the launch is a deception & hidden test    ; from [1,2]
   → the launch deceptively hides the test     ; [4]?? See below

**H:** the satellite is a ruse to hide the test
   → the launch is a ruse to hide the test     ; resolve metonymy
   → the launch is a deceptive action to hide the test   ; [2]
   → the launch deceptively hides the test     ; modal contraction

The one seemingly trivial inference which is missing is [4]:

   IF the launch is a hidden test THEN the launch hides the test
   IF X is a hidden Y THEN X hides Y

Another example is:

"The play is a hidden political commentary" → The play hides a political commentary"

**Theme: Missing inferences**

Finally, getting from "The U.S....said" (T) to "The U.S. fears" (H) is largely unwarranted from the text. I think here, "fear" is closer to fear#v4 ("be uneasy or apprehensive") than fear#v1 ("be afraid"). To make this leap we'd need to know that generally countries are uneasy/apprehensive about missile tests. DIRT doesn't have "say" → "fear", but it does have:

   IF X sees a sign of Y THEN X fears Y
   IF X expresses a concern over Y THEN X fears Y
   IF X raises a concern about Y THEN X fears Y

which are close. Also if you "try and prevent"(T) X, it means you don't want X, which often also means that you "fear"(H) X. So there's a somewhat complex but valid line of reasoning here, i.e.,:

  T says "The U.S...is working to dissuade North Korea...[from] launch[ing] a satellite",

suggesting the U.S. fears the launch.

**Theme: Modal Statements**

The nesting of modals in this example is unbelievable:

  **T:** ...the US is working to dissuade NK from going ahead with its plan to launch a satellite...

  → ...the US is working to dissuade NK from launching a satellite...

Need "going ahead with its plan to launch" → "launching", i.e.,

   "going ahead with a plan to X" → X

**Theme: Scripts**

At a deeper level, there is a familiar script at play here, and T and H are part of a common and ongoing saga. Countries want to test missiles (or other military equipment) but don't want to be seen as testing them, so may try and hide or cover up the test. Other countries are watching and are worried about deceit, and may try and put political pressure on to prevent deceipt.

**#63 YES** (BLUE said UNKNOWN)

### Full Example:

> **T:** LOSAIL, Qatar (AFP) - Australia's Casey Stoner, riding a Ducati, was confidently eyeing a third successive Qatar MotoGP victory after taking pole position in a dominant qualifying performance under the Losail desert floodlights. World champion Valentino Rossi was second fastest with his Yamaha teammate Jorge Lorenzo, who was the pole sitter in 2008, in third place. Stoner, the winner here for the last two years and the 2007 world champion, clocked 1min 55.286sec with Rossi 0.473sec behind and Lorenzo 0.497sec off the pace. Italy's Andrea Dovizioso, on a Honda, was fourth quickest.
>
> **H:** Valentino Rossi is the World Moto GP champion.

### Abbreviated example:

> **T:** ...Casey Stoner...was..eyeing a...Qatar MotoGP victory after taking pole position...World champion Valentino Rossi..
>
> **H:** Valentino Rossi is the World Moto GP champion.

### Even more abbreviated example:

> **T:** ...MotoGP...World champion Valentino Rossi..
>
> **H:** Valentino Rossi is the World Moto GP champion.

### Analysis

**Theme: Lexical variation**

Lexically, we have the "MotoGP" (T) and "Moto GP" (H) discrepancy, that need to match.

**Theme: Pragmatics, ellipsis and context**

We need to infer that "World champion Valentino" (T, sentence 2) means that Valentino is the world champion *of Moto GP*, even though this isn't stated explicitly. There is some pragmatics here, namely sentence 1 and sentence 2 both describe a race, so we assume that it's the same race (Moto GP). To make matters worse, neither sentence 1 nor sentence 2 explicitly mention a race, rather it is implicit from the ellipses in the two:

> "Stoner was eyeing a victory [in a race]"
> "Rossi was second fastest [in the race]"

Note also that it's only plausible that "World champion" refers to "Moto GP World champion" (rather than overall racing World champion, say).

**Theme: Lexicography**

Lexicographically, the name "Moto GP" sounds a bit like the name of a car or motorbike race.

**Theme: Factual world knowledge**

Other cues in the text imply the context is a motorbike race, e.g., "riding a Ducati" (people ride motorbikes), plus "Honda" and "Yamaha" are well-known brands of motorcycles.

Technically Casey Stoner could now be the new World champion if the race was over, but at time of writing he was only ahead, not the winner. "World champion" isn't qualified with "Last year's world champion" or the like -- if it were, that would add a horrible temporal aspect to the pair.

**Theme: Bracketing / Compound noun construction**

Finally the bracketing in H is horrible, as "Moto GP" is inserted between "World" and "champion":

**H:** "World Moto GP champion"

rather than "Moto GP World champion" or "World champion of the Moto GP". BLUE (mis)brackets this as:

**H:** (champion OF GP) (champion OF Moto) (champion OF World)

**#91 YES** (BLUE said UNKNOWN)

## Full example:

> **T:** Swampscott - Almost as if leading parallel careers with the 19th-century Bronte sisters are the 21st-century Ephron sisters. While Charlotte Bronte wrote "Jane Eyre," Emily wrote "Wuthering Heights" and Anne wrote "Agnes Grey." In the Ephron family, Nora wrote "Silkwood," "Heartburn," "When Harry Met Sally" and "Sleepless in Seattle," while Delia wrote "The Sisterhood of the Traveling Pants" and co-wrote "Michael" and "You've Got Mail" with Nora Â $(O a (Bll while Amy wrote six acclaimed novels. Amy Ephron's latest novel, "One Sunday Morning," is a bestseller, and film producer Jerry Bruckheimer recently bought her national bestseller, "A Cup of Tea."
>
> **H:** "Wuthering Heights" was written by Emily Bronte.

## Abbreviated example:

> **T:** ...the...Bronte sisters...While Charlotte Bronte wrote "Jane Eyre," Emily wrote "Wuthering Heights"...
>
> **H:** "Wuthering Heights" was written by Emily Bronte.

## Analysis

**Theme: Pragmatics**

This requires pragmatic knowledge about how members of a family are rendered in text, i.e., don't keep repeating the surname. This is needed to infer that "Emily" (T) means "Emily Bronte".

**#201: YES** (BLUE said UNKNOWN. This one is very difficult!)

## Full example:

| |
|---|
| **T:** Former Ontario premier Bob Rae received a boost today in his bid to win the leadership of the federal Liberal Party of Canada when rival candidate Maurizio Bevilacqua announced he was dropping out of the race in order to back Rae. "I'm convinced that Bob Rae is the best person to lead the Liberal Party to victory. He has the experience and the intelligence and the vision to lead Canada into the 21st century," Mr. Belivacqua said at a news conference, Monday. Bevilacqua is a six-term Liberal MP representing the riding of Vaughan, Ontario and was seen as a centrist on economic issues. |
| **H:** Maurizio Bevilacqua is a supporter of Bob Rae. |

## Abbreviated example:

| |
|---|
| **T:** ...Bob Rae received a boost today...when rival candidate Maurizio Bevilacqua announced he was dropping out of the race in order to back Rae. [ "I'm convinced that Bob Rae is the best person to lead the Liberal Party to victory..." Mr. Belivacqua said... ] |
| **H:** Maurizio Bevilacqua is a supporter of Bob Rae. |

## Analysis

**Theme: Scripts**

This is particularly challenging given "rival candidate" in T, which conflicts with "supporter" (H). In the scenario, Bevilacqua first was not a supporter (but a rival), but then became a supporter (after dropping out of the race). There's a common "script" here that political rivals fight but when one drops out, he/she often then backs the other.

**Theme: Modal Statements**

We also have to factor out the modal "in order to back Rae" → "back Rae".

Strictly from this sentence he didn't back Rae, he only planned to (e.g., he might have died in the meantime). However, the subsequent quote of Bevilacqua shows support - it would be *very* hard though to identify this as a supporting statement without a lot of world knowledge.

**Theme: Definitional / Word Knowledge**

We also need the rule:

> IF X backs Y THEN X is a supporter of Y.

Amazingly, this *is* in the DIRT paraphrase database, along with IF X backs Y THEN X supports Y.

**#341 YES** (BLUE said YES)

## Full example:

| |
|---|
| **T:** Over 1,600 volunteers registered to help build approximately 65,000 of the 500,000 sandbags to create dikes 20.5 feet (6.2 meters) high to protect the City of Winnipeg, Manitoba in the war against the Red River of the North flood. 700 volunteers answered at the rural municipality of St. Andrews alone. Once sandbags are filled for West St. Paul, St. Andrews, and Selkirk, then frozen culverts must be cleared. The height of the river is expected to be Thursday, and predictions are that it will be less than Flood of the Century of 1997. There is no precipitation in the forecast, and snow in the province should be melted by the end of the week. |
| **H:** Red River will overflow its banks. |

## Abbreviated example:

| |
|---|
| **T:** ...volunteers registered to help build...to create dikes...to protect...Winnipeg... in the war against the Red River...flood. |
| **H:** Red River will overflow its banks. |

## Even more abbreviated example:

| |
|---|
| **T:** ...the Red River...flood.... |
| **H:** Red River will overflow its banks. |

## Analysis

**Theme: Reasoning with World/Lexical knowledge**

This example requires linking "overflow its banks" with "flood" - it seems like it should be easy, but it's surprisingly difficult. WordNet gives us:

> flood#n1: the rising of a body of water and its overflowing onto normally dry land

This gives us "flood" (T) → "overflow" (H) but we'd also need "Red River" (H) and "its banks" (H).

For "Red River", we need to identify that "Red River" (T) is coreferential with "the body of water" (WordNet definition), however the WordNet definition gives us no clues as to the possible syntactic relationship between "flood" and the "body of water". We could assume NN compounding is one possibility for such a relationship.

For "its banks", the world knowledge rule we need is:

> IF a river floods THEN the river overflows its banks

WordNet also gives us:

> bank#n2: sloping land (especially the slope beside a body of water)
> overflow#v1: flow or run over (a limit or brim)

Also that {flood,overflow} are synonyms:

> {flood#n4,overflow#n2,outpouring#n5}: a large flow

> flood#v1 (6) deluge, flood, inundate, swamp -- (fill quickly beyond capacity; as with a liquid; "the basement was inundated after the storm"; "The images flooded his mind")

> flood#v2 (3) flood -- (cover with liquid, usually water; "The swollen river flooded the village"; "The broken vein had flooded blood in her eyes")

None of these quite embody the world knowledge that a river flooding involves the river overflowing its banks, although it gets "tantalizingly close". flood#n1 gives "overflowing onto dry land", bank#n2 gives us "sloping land by a body of water", but nothing that the bank is between the river and dry land. We need knowledge that the bank is a kind of barrier on the river.

Looking at other resources for this missing knowledge, namely PowerSet Factz and DART, they which contain triples (i.e., observations in text) of:

> "floods can overflow banks"

and:

> "rivers can overflow banks"

However, the PowerSet Factz for "rivers can overflow banks" occur in 3 articles about hurricanes and typhoons. (I was hoping they'd be articles about floods :-( ).

**Theme: Bracketing/Parsing**

"the Red River of the North flood" is difficult to parse. Note the bracketing is different in the two below, differing in only a single letter's case:

> "the Red River of the North flood" → "(the Red River of the North) flood"
> "the Red River of the North Flood" → "the Red River of (the North Flood)"

**#343 YES** (BLUE said UNKNOWN)

## Full example:

> **T:** USC Trojans running back Reggie Bush has decided to enter the NFL draft and forfeit his final year of college football eligibility. The Heisman Trophy winner is the favorite to be the top pick in the draft, in which case he would play for the Houston Texans. Bush rushed for 1,740 yards during his final college season, and led the Trojans to the Rose Bowl. Quarterbacks Marcus Vick of the Virginia Tech Hokies and Vince Young of the Texas Longhorns are also likely to among the top draft picks this year.
>
> **H:** Reggie Bush is the recipient of the Heisman Trophy.

## Abbreviated example:

> **T:** ...Reggie Bush...The Heisman Trophy winner...
>
> **H:** Reggie Bush is the recipient of the Heisman Trophy.

## Analysis

**Theme: Coreference**

This is again a common pattern of complex coreference, where the reference is a new description ("the winner") of the referent ("Reggie Bush") rather than a repeat of the referent's name. This is a somewhat stylistic convention in authoring to make text less boring (but harder for machines!).

**Theme: Reasoning with World/Lexical knowledge**

We also need to get from "winner" to "recipient". WordNet doesn't have a relationship, but DIRT has exactly the rule we need:

IF X is the winner of Y THEN X is the receipient of Y

Impressive!

Strictly this isn't deductive (you can win something without receiving it, e.g., an election, competition), but usually it is (when you win a tangible reward/trophy/prize, or something accompanied with a reward/trophy/prize).

**#348 YES** (BLUE said UNKNOWN)

## Full example:

| |
|---|
| **T:** The alleged mastermind behind the London bombings was reported captured in Cairo, Egypt last week. Police believe that a U.S. trained chemist, Magdi Asdi el-Nashar, 33, helped build the bombs that killed over 50 people. Mr. el-Nashar, who has a PhD from Leeds University, left England two weeks before the bombings. After the London bombings, British authorities initiated a worldwide manhunt that found him in Cairo. State security officials reported they have begun questioning el-Nashar with British agents in attendance. |
| **H:** Cairo is situated in Egypt. |

## Abbreviated example:

| |
|---|
| **T:** ...Cairo, Egypt... |
| **H:** Cairo is situated in Egypt. |

## Analysis

**Theme: Appositive interpretation, geographic knowledge**

The apposition "Cairo, Egypt" commonly means X is geographically in Y i.e., the semantic relation is something like "location" or "located in".

WordNet gives us {"locate","situate"} are synonyms.

This example requires world knowledge that Cairo and Egypt are places (which WordNet can provide).

**#532 YES** (BLUE said YES). This example is easy.

## Full example:

| |
|---|
| **T:** Sheikh Hasina (Shekh Hasina Oajed) was the Prime Minister of Bangladesh from 1996 to 2001. She has been the President of the Awami League, a major political party in Bangladesh, since 1981. She was the second female prime minister of Bangladesh. Bangladesh, officially the People's Republic of Bangladesh ( Gônoprojatontri Bangladesh), is a country in South Asia. It is bordered by India on all sides except for a small border with Burma (Myanmar) to the far southeast and by the Bay of Bengal to the south. Together with the Indian state of West Bengal, it makes up the ethno-linguistic region of Bengal. |
| **H:** Sheikh Hasina was Bangladesh's Prime Minister. |

## Abbreviated example:

| |
|---|
| **T:** Sheikh Hasina (Shekh Hasina Oajed) was the Prime Minister of Bangladesh... |
| **H:** Sheikh Hasina was Bangladesh's Prime Minister. |

Simple syntactic manipulation will solve this one.